\RequirePackage{fix-cm}
\documentclass[smallextended]{svjour3}       
\smartqed  
\usepackage{graphicx}
\usepackage{amssymb}
\usepackage{latexsym}
\usepackage{float}
\usepackage{hyperref}
\usepackage[
  separate-uncertainty = true,
  multi-part-units = repeat
]{siunitx}

\begin{document}

\title{Transfer-Learning-Aware Neuro-Evolution for Diseases Detection in Chest X-Ray Images
}


\author{Albert Susanto         \and
        Herman \and
        Tjeng Wawan Cenggoro \and
        Suharjito \and
        Bens Pardamean
}


\institute{Albert Susanto \at
              \email{albert.susanto001@binus.ac.id}           
\and
           Herman \at
              \email{herman005@binus.ac.id}
              \and
           Tjeng Wawan Cenggoro \at
              \email{wcenggoro@binus.edu}
              \and
           Suharjito \at
              \email{suharjito@binus.edu}
              \and
           Bens Pardamean \at
              \email{bpardamean@binus.edu}
}

\date{Received: date / Accepted: date}

\maketitle

\begin{abstract}
The neural network needs excessive costs of time because of the complexity of architecture when trained on images. Transfer learning and fine-tuning can help improve time and cost efficiency when training a neural network. Yet, Transfer learning and fine-tuning needs a lot of experiment to try with. Therefore, a method to find the best architecture for transfer learning and fine-tuning is needed. To overcome this problem, neuro-evolution using a genetic algorithm can be used to find the best architecture for transfer learning. To check the performance of this study, dataset ChestX-Ray 14 and DenseNet-121 as a base neural network model are used. This study used the AUC score, differences in execution time for training, and McNemar’s test to the significance test. In terms of result, this study got a 5 \% difference in the AUC score, 3 \% faster in terms of execution time, and significance in most of the disease detection. Finally, this study gives a concrete summary of how neuro-evolution transfer learning can help in terms of transfer learning and fine-tuning.
\end{abstract}

\section{Introduction}
Lungs are an important organ for humans. Some of the diseases in the lungs like atelectasis, pleuritis, pneumonia can be detected by using a chest X-ray. Although, there are many diseases in the lungs. The deadliest is pneumonia \cite{rajpurkar2017chexnet}. Every year there are 1 million adults that are suffering from pneumonia and there are 50,000 casualties every year in the United States for every year \cite{rajpurkar2017chexnet}. Even though, there is already a medicine for pneumonia treatment diagnosing method for pneumonia is still needed. The best method right now for pneumonia detection in chest X-rays. However, there is a lack of expert radiologists that makes detecting method using chest X-rays become a challenging task. Diseases detection made an improvement when there is a group of researchers develop a dataset called ChestX-ray 14 dataset \cite{wang2017chestx}. With the assistance of ChestX-ray 14, many researchers try to use neural networks for disease detection.

Convolutional Neural Network (CNN) has been a solution for image detection \cite{lecun1989backpropagation,krizhevsky2012imagenet,hu2018squeeze}. In the previous work of the researchers, they made a detecting method using a neural network that called CheXNet that consists of 121-layer of CNN with Adam optimizer \cite{rajpurkar2017chexnet}. In that research, they used CheXNet, DenseNet-121 neural network with transfer learning and fine-tuning from ImageNet. CheXNet also used ChestX-ray 14 dataset for training. Even without any modification, they can surpass the latest state of the art for detecting pneumonia, which is radiologist performance in ChestX-ray14 datasets. Even though CheXNet already surpasses the latest state of the art, there are other researchers that beat CheXNet using DenseNet-121 and inject the Squeeze and Excitation layer between the transition layer \cite{yan2018weakly}. When trained on images, the neural network needs a lot of time because of the complexity of architecture \cite{pardamean2018transfer}. Therefore, transfer learning and fine-tuning can help improve time and cost efficiency when training a neural network \cite{ghazi2017plant}.

Transfer learning is the method that let neural network to learn with the knowledge of previous work. Basically, this method is used to improve time efficiency when training neural networks \cite{ghazi2017plant}.  Transfer learning can be applied by transfer all the weight from the previous trained neural network to a new neural network. Neural networks nowadays share something like Gabor filter or color blobs in the first few layers when trained on images. That means, the first few layers can be frozen and then train the rest of the layer or remove some layers that didn’t correlate with the datasets and tasks, this is called the fine-tuning method \cite{yosinski2014transferable}. Therefore, time efficiency can be improved when training and testing. With the same accuracy, the shorter the neural network, the better in terms of execution time. 

Transfer learning and fine-tuning needs a lot of experiment to try with. In the other paper, there are researchers that use transfer learning and fine-tuning neural network to specify different task \cite{kim2018artificial,huang2016unified,oquab2014learning,too2019comparative,shin2016deep}. Afterward, transfer learning and fine-tuning can only be achieved by trying and repeating all process and get the best condition for the current datasets. Approximately, transfer learning and fine-tuning need an average of 2 hours per process for 20 epochs training and all of it has been done in a manual way \cite{pardamean2018transfer}. 

Therefore, a method that let machine discover new network architectures is needed. One of the methods is called neuro-evolution that uses a genetic algorithm \cite{miller1989designing}. Previous research shows that neuro-evolution discovers novel architecture with state-of-the-art performance \cite{real2019regularized}. Afterward, there is also paperwork that researches the discovery of new architecture in CIFAR-10 datasets and then tests it for image classification on different datasets using a genetic algorithm \cite{zoph2018learning}. The purpose of neuro-evolution by previous researchers is to discover new architecture that is different from the current neural network. Because of that, they cannot apply transfer learning and fine-tuning. Therefore, neuro-evolution transfer learning was proposed, a method for discovering crucial layers of the DenseNet-121 + Squeeze and Excitation (SE) layer in transfer learning and fine-tuning using a genetic algorithm. This method is not used to discover new architecture. Instead, this method found the best CNN architecture for transfer learning and its hyperparameter by searching using the genetic algorithm. Previously mentioned method called as transfer-learning aware neuro-evolution.

Our contribution is summarized as follows:
\begin{itemize}
 \item Develop a transfer-learning aware neuro-evolution for discovering new transfer learning architecture in DenseNet-121 + SE layer using a genetic algorithm.
 \item Improving AUC score in chest X-ray diseases detection using transfer-learning aware neuro-evolution for discovering new transfer learning architecture in DenseNet-121 + SE layer using a genetic algorithm.
 \item Improving time efficiency in chest X-ray diseases detection using transfer-learning aware neuro-evolution for discovering new transfer learning architecture in DenseNet-121 + SE layer using a genetic algorithm.
\end{itemize}

\section{Literature study}
\subsection{Densely Connected Convolutional Network}
Densely Connected Convolutional Network (DenseNet) is a newly implemented model of CNN \cite{krizhevsky2012imagenet}. DenseNet consists of some dense block layer as shown in Figure 1. Each block in DenseNet consists of a different number of layers depends on the type of DenseNet. In DenseNet-121 there are 4 blocks that consist of 6, 12, 24, 16 layers consecutively.

\begin{figure}
  \includegraphics[scale=.5]{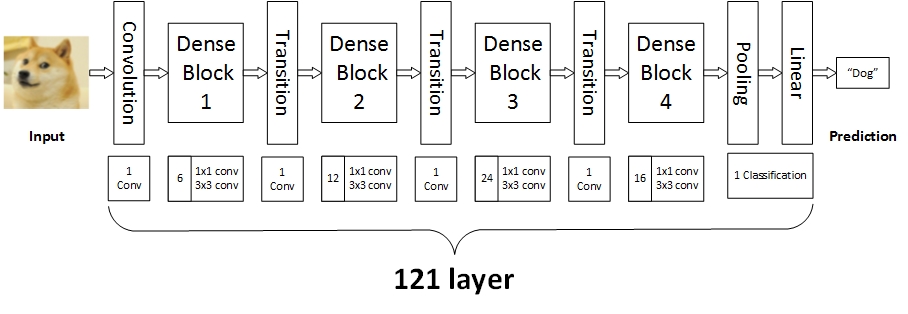}
  \caption{DenseNet architecture \cite{rajpurkar2017chexnet}}
\end{figure}

DenseNet layers in every block are connected in a feed-forward fashion. This means, in every forward process every layer in the blocks got information about weight from the past layers. Afterward, in every backward (Backpropagation) process every layer in the blocks got loss and gradient information from the past layers. This connection can be seen in Figure 2.

\begin{figure}

\includegraphics[scale=.75]{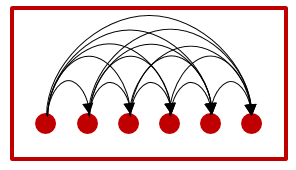}%
\caption{The connection between layers in blocks of DenseNet \cite{huang2017densely}}
\end{figure}

DenseNet has many of advantages like all layer in block have direct access to gradient of the loss function (vanishing gradient problem), all features that learned by the first layer can be accessed by throughout all layers in the same block (robust feature propagation), and all the subsequent layers have all of the information from all of the past layers (reduce the number of parameters).

\begin{figure}[b]

  \includegraphics[scale=.45]{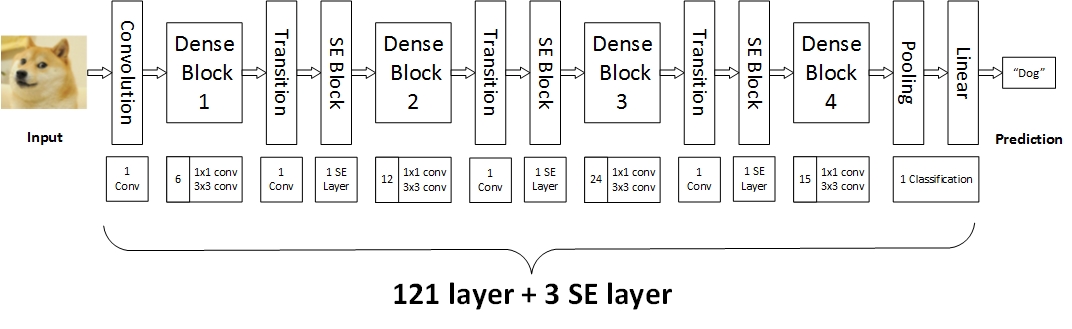}
  \caption{Densenet + SE Layer Architecture \cite{yan2018weakly}}
\end{figure}

\subsection{Squeeze and Excitation Layer}
The squeeze and excitation layer is a computational unit that injected after the convolutional layer. SE layer has a purpose to recalibrate feature weight with global channel-wise attention \cite{hu2018squeeze}.

Squeeze used to capture network weights of its channel and then extract the weight using average pooling with reduction ratio. Afterward, excitation are used to recalibrate feature maps using the Relu activation method. The result of this operation needs to multiply into neural network feature maps.

The squeeze and excitation layer uses a simple mathematical equation. Therefore, squeeze and excitation layer is easy to be implemented in all neural networks.

The last layer from the transition layer represented by are calculated with element-dot wise with the squeeze and excitation layer. The result of this calculation represent as input in the new dense block layer. The DenseNet-121 + SE layer architecture can be seen in Figure 3.
\begin{figure}[b]

  \includegraphics[scale=.75]{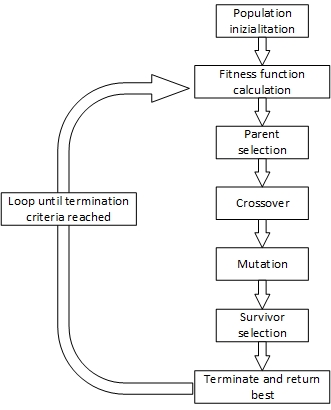}
  \caption{Genetic Algorithm Architecture \cite{goldberg1991comparative}}
\end{figure}

\subsection{Genetic Algorithm}
Genetic algorithm is one of the oldest methods that can be used to solve the optimization problem \cite{goldberg1991comparative}. The genetic algorithm consists of several steps that can be seen in Figure 4. The genetic algorithm was inspired by Darwin's theory of biological evolution. The genetic algorithm starts with generating some cells that are called a chromosome. And that group of chromosomes are called a population that needs to be evolved. After that, every chromosome fitness value are calculated based on fitness rules. In every generation, every chromosome reproduce in the crossover and mutation phase. Crossover is a phase where every cell in a chromosome can be traded. Meanwhile, in mutation, every cell has a chance to change a little bit based on mutation rules too. After all of that, not every chromosome can live and proceed to the next generation. The only chromosome that fits with the criteria can proceed to the next generation. And this process repeat until the criteria are reached.

\subsection{Neural Network on Chest X-ray Diseases Detection}
The most powerful method used for detecting lung diseases was detecting via a chest X-ray. Chest X-ray then are analyzed by expert radiologists. However, due to a lack of expert radiologists, researchers start to make a deep learning model that can be used for detecting pneumonia. With this kind of motivation, researchers start to make a database for chest X-ray that called ChestX-ray 8 database. This database consists of 112,120 X-ray images from 30,805 unique patients that were created by \cite{wang2017chestx}. This database labeled with automatic extraction methods based on radiology reports. After being labeled, this database was group into 14 different thoracic pathologies. This database later called as the ChestX-ray 14 database (National Institute of Health Chest X-ray datasets).

After that, researchers start to implement a model of CNN which is called DenseNet-121. In the previous work, DenseNet-121 was a popular neural network that been used to solve many problems in diseases detection \cite{rajpurkar2017chexnet,yan2018weakly,guan2018multi,gottapu2018densenet,que2018cardioxnet,antindetecting}. DenseNet-121 with 121 layers can beat VGG-Net with 16 layers and ResNet with 50, 101, 152 layers in the same condition \cite{too2019comparative}. CheXNet using Densenet with 121 layers and Adam optimization \cite{kingma2014adam} as its core. In the research, CheXNet can be more accurate than radiologist experts in pneumonia detection. CheXNet got a 0.435 F1 score and radiologist experts only got an average of 0.387. Besides that, CheXNet also makes measurement using the AUC (Area Under Receiver Operating Characteristic Curve) score. AUC range from 0 to 1. Therefore, if all predictions are 100 \% correct, it has an AUC score of 1 and CheXNet got an average AUC score of 0.84138 to predict 14 types of lung diseases. 

Until now, there are already some solutions using neural networks for detecting diseases from chest X-ray images. The most popular neural network that been used is ResNet and DenseNet-121 \cite{que2018cardioxnet, li2018thoracic}. After using ResNet or DenseNet-121 neural network, there is some method that tries by the researcher like: try to augment the datasets from ChestX-Ray 14 \cite{antindetecting}, try using attention mechanism at the end of the neural network \cite{wang2018chestnet} or adding SE layer in every transition layer \cite{yan2018weakly}. After all of that, CheXNet got a slightly better result than any other method as seen in Table 1. Even though CheXNet got a slightly better result, by adding the SE layer on DenseNet-121 there is an improvement by almost 2 \% inaccuracy according to the SE layer paper \cite{yan2018weakly}.  Therefore, the DenseNet-121 + SE layer was used as the base neural network model.

\begin{table*}[t]
\caption{\label{tab1}State of the art for diseases detection in chest X-ray images.}

\begin{tabular}{|p{1cm}|p{4cm}|p{2cm}|p{2cm}|p{1cm}|}
\hline
Author  & Algorithm used & Dataset & Transfer learning method & Average AUC score \\ \hline
\cite{antindetecting}    & Augment datasets and using DenseNet-121 & ChestX-ray 14 & Pre-trained on ImageNet & 0.609 \\ \hline
\cite{wang2018chestnet}    & Using ResNet 152 + attention mechanism & ChestX-ray 14 & Pre-trained on ImageNet & 0.781 \\ \hline
\cite{guan2018multi}    & Using CNN + attention mechanism & ChestX-ray 14 & - & 0.816 \\ \hline
\textbf{\cite{yan2018weakly}}     & \textbf{Add SE layer between transition every layer in DenseNet-121} & \textbf{ChestX-ray 14} & \textbf{Pre-trained on ImageNet} & \textbf{0.830} \\ \hline
\textbf{\cite{rajpurkar2017chexnet}}    & \textbf{Using DenseNet-121 and Adam optimizer} & \textbf{ChestX-ray 14} & \textbf{Pre-trained on ImageNet} & \textbf{0.841} \\ \hline
\cite{wang2018chestnet}    & Using ResNet 152 + attention mechanism & ChestX-ray 14 & Pre-trained on ImageNet & 0.781 \\ \hline
\end{tabular}
\end{table*}

\subsection{Neuro-evolution Algorithm}
The neuro-evolution algorithm already used by Google researchers to solve a problem. Google researchers use Neuro-evolution with a genetic algorithm to discover new deep learning architecture. This deep learning architecture learns from the CIFAR-10 dataset. Form the report, this new deep learning architecture can outmatch the current state of the art for CIFAR-10 dataset in 83.9 \% accuracy for top-1 accuracy and 96.6 \% accuracy for top-5 accuracy \cite{real2019regularized}.

There is paperwork where a genetic algorithm combined with transfer learning \cite{Zech2018}. In this paper, the researcher trains the machine learning with genetic algorithm and then transfer 30 \% of the best population as a starter to predict different task this method is called genetic transfer learning. Genetic transfer learning is kind of different from our approach, where our approach wants to find the most suitable transfer learning architecture for CheXNet instead of transfer learning the genetic from before.

Time for training and discovering a new model becomes important when train neural networks. Our approach builds based on the previous DenseNet-121 + SE layer model. The genetic algorithm mutate chromosomes and try each of the combinations of the DenseNet-121 + SE layer model. Further, the current state of the art only using the default configuration for chest X-ray diseases detection using DenseNet-121 + SE layer model and Adam optimization. Even though there are many hyperparameters that can be experimented with. The initial weights are filled from the pre-trained ImageNet and optimized using the SGD+momentum optimization method in training. Therefore, the genetic algorithm was used to discover the fittest architecture for transfer learning in DenseNet-121 + SE layer model. This algorithm combines each of the hyperparameters in the method that called crossover. Every hyperparameter becomes the variable for the crossover and mutation, such as learning rate, number of included layers, number of training layers, etc.

\section{3. Research method}
The genetic algorithm was used to get the most suitable transfer learning architecture for chest X-ray diseases detection. The genetic algorithm itself consists of several steps: population initialization, fitness function calculation, parent selection, crossover, mutation, survivor selection, termination.

The population initialization in the genetic algorithm was initialized based on previous research by \cite{pardamean2018transfer}. Thus, every chromosome has a cell that contained:
\begin{itemize}
 \item Included layers (Number of used layers in most suitable blocks)
A random number between 1 – 58 
Included layers are the number of accumulated layers in every block

 \item Frozen layers (Number of freeze layers in included layers)
A random number between 0 – 2nd block of DenseNet-121. 
Freeze layers began from the input layer until the number of accumulated layers from the frozen layers parameter.

 \item Learning rates
Random number from 0.1, 0.01, 0.001, 0.0001, 0.00001, 0.000001.

\item Dropout
Random number from 0.1, 0.2, 0.3, 0.4, 0.5, 0.6, 0.7, 0.8, 0.9.
\end{itemize}

The chromosome is illustrated in Figure 5 and every generation composed of 10 chromosomes.

\begin{figure}
    
    \includegraphics[scale=.75]{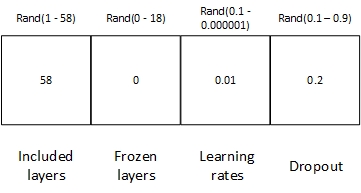}
    \caption{Chromosome Initialization}
\end{figure}{}

For the fitness function calculation, each chromosome is ranked by the value of average binary cross-entropy loss from 5 training epochs as recommended in \cite{cs231n}. The initial weight is loaded from pre-trained DenseNet-121 that has been trained in ImageNet before. The implementation of CheXNet was acquired from \cite{Zech2018} which, already been optimized using SGD + Momentum optimization instead of Adam optimization. Afterward, the SE layer model code was implemented to get a DenseNet-121 + SE layer model. Each average loss was multiplied with -1 to make sure that the smallest loss has the highest rank. Every chromosome fitness value was kept to avoid the calculation of the same chromosome.

Next, in the parent selection, the tournament selection method was used, where the method pick 2 random chromosomes in population and let the bigger fitness value win and become one of the parents. After the first parent was acquired, the method was run 1 more time to get the second parent.

Afterward, there is a crossover phase, where parents’ chromosomes were combined to get a newly born child chromosome. For this phase, the uniform crossover method was chosen, every parent trade its chromosome. There is a 50 -50 chance every cell in the chromosome traded.

And then after that, the mutation is a slight tweak in the genetic algorithm. It gives a genetic algorithm to discover new search space. In this step, every cell value in a chromosome has a slight chance to be added or subtracted. the chromosome was mutated with rules:

\begin{itemize}
 \item Included layers are mutated by \SI{ \pm 5} layers.
 \item Frozen layers are mutated by \SI{ \pm 5} layers.
 \item Learning rates are mutated by multiple or division by 10.
 \item Dropout are mutated by \SI{ \pm 0,1}.
\end{itemize}

The chance should be tiny, so the genetic algorithm would not be a random search process. A 10 \% chance was used by the genetic algorithm to perform mutation. The survivor selection step is a straightforward step. 

Where every generation have 2 parents and a new population that is built by crossover and mutation step. Typically, the old chromosome can be in next-generation too if the crossover and mutation step fails to perform. As the termination process, the process from fitness function calculation to survivor selection was repeated until the termination criteria are reached, which is 10 generations or there is a small difference in average fitness function per generation, which is 0,001.

And then after all of the steps before, the next step was to run the genetic algorithm method. The process implemented in PyTorch using NVIDIA Tesla P100 GPU in Bina Nusantara University AI R\&D Center. After the genetic algorithm was converged, the best chromosome that has the smallest loss was picked as the best chromosome. Thus, the best chromosome was re-once more for 100 epochs and then comparing the testing result prediction with the DenseNet-121 + SE layer that already trained before.

\begin{figure*}[!t]

\includegraphics[scale=.75]{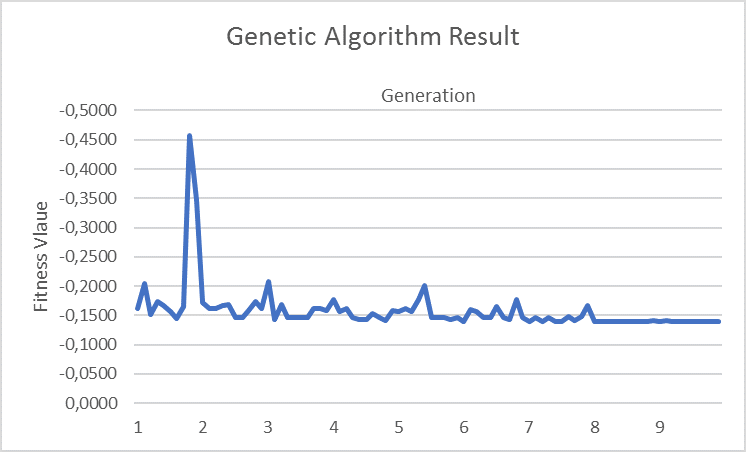}
\caption{Genetic Algorithm Result}
\end{figure*}

\begin{figure*}[b]

\includegraphics[scale=.45]{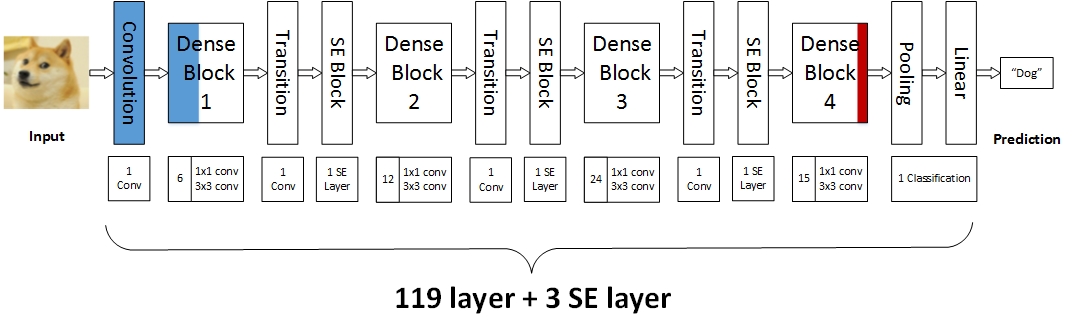}
\caption{Transfer-Learning-Aware Neuro-Evolution Model}
\end{figure*}

\begin{table*}[t]
\caption{\label{tab1}Result research in AUC score}

\begin{tabular}{|p{3cm}|p{3cm}|p{3cm}|p{3cm}|}
\hline
Diseases  & \cite{rajpurkar2017chexnet} & \cite{yan2018weakly} & Transfer-Learning Aware Neuro-Evolution Model \\ \hline
Atelectasis & 0.735377 & 0.730659 & \textbf{0.766682}  \\ \hline
Cardiomegaly &	0.848622 &	0.851578 &	\textbf{0.884737}\\ \hline
Consolidation &	0.717086 &	0.715865 &	\textbf{0.741742}\\ \hline
Edema &	0.827011 &	0.823502 &	\textbf{0.839080}\\ \hline
Effusion &	0.800781 &	0.802321 &\textbf{0.829629}\\ \hline
Emphysema &	0.779976 &	0.772761 &	\textbf{0.890848}\\ \hline
Fibrosis &	0.747934 &	0.749378 &	\textbf{0.812461}\\ \hline
Hernia &	0.688933 &	0.716595 &	\textbf{0.770063}\\ \hline
Infiltration &	0.672221 &	0.674133 & \textbf{0.686300}\\ \hline
Mass &	0.695092 &	0.706002 &	\textbf{0.810984}\\ \hline
Nodule &	0.657002 &	0.660613 &	\textbf{0.733663}\\ \hline
Pleural Thickening &	0.719719 &	0.718566 & \textbf{0.766746}\\ \hline
Pneumonia &	0.688964 &	0.678925 &	\textbf{0.717143}\\ \hline
Pneumothorax &	0.800578 &	0.796132 &	\textbf{0.855771}\\ \hline
Average AUC Score &	0.741378 &	0.742645 &	\textbf{0.793275}\\ \hline
\end{tabular}
\end{table*}

In order to prove the contribution, comparing the AUC score was used to evaluate our neuro-evolution transfer learning model with the previous work \cite{rajpurkar2017chexnet}, \cite{yan2018weakly}. AUC is a method that is very popular on second after accuracy. AUC method calculates how well did our model performs and makes the prediction. And total execution time in training and testing were used to evaluate our transfer-learning aware neuro-evolution model with the previous work. Afterward, McNemar’s test was used to prove that our transfer-learning aware neuro-evolution model was truly significant. The McNemar’s test was run using R \cite{rDasar,rLanjut}.

\section{Results \& discussion}
The dataset used for this study is from the National Institute of Health (NIH) released by \cite{wang2017chestx}. Which contains 112,120 frontal-view of X-ray images from 30,805 unique patients. The ChestX-ray 14 datasets were labeled for 14 different thoracic pathologies from radiology reports using automatic extraction methods. This dataset is called ChestX-ray 14 dataset. The division of the dataset using the same testing configuration as the ChestX-ray14 dataset \cite{wang2017chestx}, which is 25.596 (22 \%) images for testing. The rest of the images are split randomly with 70 \% data for training, and 8 \% for validation. The image resolution from the NIH is 1024 x 1024 pixels. Therefore, the images should be shrunk into 224 x 224 pixels as input for DenseNet-121. 

\begin{table*}[b]
\caption{\label{tab1}Total execution time in training}

\begin{tabular}{|p{3cm}|p{3cm}|p{3cm}|p{3cm}|}
\hline
Diseases  & \cite{rajpurkar2017chexnet} & \cite{yan2018weakly} & Transfer-Learning Aware Neuro-Evolution Model \\ \hline
Execution Time Training &	\textbf{1279m 48s} &	1414m 53s &	1362m 29s\\ \hline
\end{tabular}
\end{table*}

The convergence of the genetic algorithm can be seen when the fitness value of each chromosome hits an elbow point. The elbow point is a point where the fitness value is not improving. As seen in Figure 6, the convergence of the transfer-learning aware neuro-evolution model genetic algorithm occurred from generation 8. The evolutionary stop in generation 9 because the difference between average fitness value per generation is $9,5259 \times 10^{-5}$. Therefore, the combination hyperparameters that used in the transfer-learning aware neuro-evolution method neural network are the best generation starting from generation 9, which is 57 used layers; 2 freeze layers; 0,1 learning rate; and 0,1 dropouts as seen in Figure 7. The transfer-learning aware neuro-evolution model architecture used a total of 119 layers of DenseNet-121 + 3 SE layers. The blue color indicates a total layer that affected by freeze layers configuration and the red color indicates a total layer that removed by used layers configuration.

\begin{table*}[!t]
\caption{\label{tab1}Significance test using McNemar’s}

\begin{tabular}{|p{3cm}|p{2cm}|p{2cm}|p{2cm}|p{2cm}|}
\hline
Diseases  & \cite{rajpurkar2017chexnet} & Significance & \cite{yan2018weakly} & Significance \\ \hline
Atelectasis &	$<$ 2.2e-16 &	\textbf{TRUE} &	$<$ 2.2e-16 &	\textbf{TRUE}\\ \hline
Cardiomegaly &	$<$ 2.2e-16 &	\textbf{TRUE} &	$<$ 2.2e-16 &	\textbf{TRUE}\\ \hline
Consolidation &	- &	FALSE &	- &	FALSE\\ \hline
Edema &	6.769e-10 &	\textbf{TRUE} &	3.435e-05 &	\textbf{TRUE}\\ \hline
Effusion &	$<$ 2.2e-16 &	\textbf{TRUE} &	4.609e-12 &	\textbf{TRUE}\\ \hline
Emphysema &	$<$ 2.2e-16 &	\textbf{TRUE} &	$<$ 2.2e-16 &	\textbf{TRUE}\\ \hline
Fibrosis &	1.814e-05 &	\textbf{TRUE} &	1.944e-4 &	\textbf{TRUE}\\ \hline
Hernia &	1e0 &	FALSE &	3.948e-3 &	\textbf{TRUE}\\ \hline
Infiltration &	$<$ 2.2e-16 &	\textbf{TRUE} &	$<$ 2.2e-16 &	\textbf{TRUE}\\ \hline
Mass &	$<$ 2.2e-16 &	\textbf{TRUE} &	$<$ 2.2e-16 &	\textbf{TRUE}\\ \hline
Nodule &	$<$ 2.2e-16 &	\textbf{TRUE} &	$<$ 2.2e-16 &	\textbf{TRUE}\\ \hline
Pleural Thickening &	1.504e-10 &	\textbf{TRUE} &	1.504e-10 &	\textbf{TRUE}\\ \hline
Pneumonia &	4.795e-1 &	FALSE &	4.795e-1 &	FALSE\\ \hline
Pneumothorax &	$<$ 2.2e-16 &	\textbf{TRUE} &	$<$ 2.2e-16 &	\textbf{TRUE}\\ \hline
\end{tabular}
\end{table*}

The result of this study can be seen in Table 2. There is an improvement in the AUC score. However, the Densenet-121 + SE Layer model \cite{yan2018weakly} and the transfer-learning aware neuro-evolution method has the same neural network model. The transfer-learning aware neuro-evolution method got a 0.793275 average AUC score, which is 5 \% higher than the previous work. Furthermore, there is an improvement in the detection of every disease. This proofs that the discovered hyperparameter setting is better than the setting used by the previous work.

The comparison of execution time in training between all compared models can be seen in Table 3. CheXNet \cite{rajpurkar2017chexnet} has the fastest training execution time from all of the compared models. This is happening because of CheXNet didn’t add the SE layer between the transition layer of the neural network model. Thus, CheXNet needs a lesser computational process than other neural network models. In comparison between the transfer-learning aware neuro-evolution model and DenseNet-121 + SE layer model, the transfer-learning aware neuro-evolution model got a better result of 52 minutes and 24 seconds or 3 \% faster. The training time difference between these two models is caused by the averted weights-update computation in the frozen layer during training.

Lastly, the significance level in McNemar’s Test can be measured if the p-value is less than 0.05. For consolidation disease, the significance level of McNemar’s cannot be calculated because there is no change in prediction between all models. The McNemar’s test did not get fully significance in hernia and pneumonia diseases because there is only a slight difference between neuro-evolution transfer learning model and the previous model.

The neuro-evolution transfer learning model got 11 and 12 significance diseases detection respectively against CheXNet and DenseNet-121 + SE layer model as seen in Table 4. Therefore, this research accepts the hypothesis for 11 out of 14 diseases against CheXNet and accepts the hypothesis for 12 out of 14 diseases against DenseNet-121 + SE layer model. Furthermore, this research did not try on the other model neural network due to lack of resources and time.

\section{Conclusion}
A neuro-evolution algorithm with a genetic algorithm is a powerful tool or method to find the optimized transfer learning architecture. This research found that not every layer in the model is important to train a certain task or dataset. This research also found that the neuro-evolution algorithm with a genetic algorithm can be implemented in DenseNet-121 + SE layer architecture to find the optimized transfer learning architecture for 11 diseases. Neuro-evolution algorithm with genetic algorithm also can help improving 3 \% faster in time efficiency and 5 \% in AUC score from DenseNet-121 + 3 SE layer model. 

For future research, this research use four cell in every chromosome. Therefore, adding a different optimization method for training or learning rate decay epoch in every chromosome can be used for the next research. Furthermore, a different optimization technique can be implemented in the neuro-evolution algorithm since this research only use genetic algorithm.

\section*{Acknowledgments}
The NVIDIA P100 GPU used in this work is supported by NVIDIA – BINUS AI R\&D Center.

\section*{Conflict of interest}

The authors declare that they have no conflict of interest.

%
%

\end{document}